\documentclass[conference]{IEEEtran}
\pdfoutput=1 
\IEEEoverridecommandlockouts
\usepackage{cite}
\usepackage{amsmath,amssymb,amsfonts}
\usepackage{algorithmic}
\usepackage{graphicx}
\usepackage{textcomp}
\usepackage{xcolor}
\usepackage[utf8]{inputenc}
\usepackage{url}
\usepackage{flushend}
\def\BibTeX{{\rm B\kern-.05em{\sc i\kern-.025em b}\kern-.08em
    T\kern-.1667em\lower.7ex\hbox{E}\kern-.125emX}}
    
\DeclareMathOperator*{\argmin}{argmin}    
    
\begin{document}

\title{Absolute Human Pose Estimation with Depth Prediction Network
\thanks{This work was completed in the ELTE Institutional Excellence Program 
(1783-3/2018/FEKUTSRAT) supported by the Hungarian Ministry of Human Capacities. A.L. was supported by part through grant EFOP-3.6.3-VEKOP-16-2017-00002. }
}

\author{\IEEEauthorblockN{M\'arton V\'eges and András Lőrincz}
\IEEEauthorblockA{\textit{Department of Software Technology and Methodology} \\
\textit{Eötvös Loránd University}\\
Budapest, Hungary}
}

\maketitle

\begin{abstract}
The common approach to 3D human pose estimation is predicting the body joint coordinates relative to the hip. This works well for a single person but is insufficient in the case of multiple interacting people. Methods predicting absolute coordinates first estimate a root-relative pose then calculate the translation via a secondary optimization task. We propose a neural network that predicts joints in a camera centered coordinate system instead of a root-relative one. Unlike previous methods, our network works in a single step without any post-processing. Our network beats previous methods on the MuPoTS-3D dataset and achieves state-of-the-art results.
\end{abstract}


\section{Introduction}
Human pose estimation has received a lot of attention recently due to its various potential applications, for example in augmented reality, sports analytics or rehabilitation. While 2D pose estimators have reached good results \cite{openpose,stacked_hourglass,alphapose}, 3D pose prediction still has areas to improve.

One difficulty of the problem comes from the fact that fully annotated 3D databases are hard to create. To create accurate measurements, special equipment with multiple cameras, depth sensors and adequate synchronization are needed. There are only a few in-the-wild datasets, most databases were created in a studio. Also,  monocular 3D pose estimation is inherently ambiguous. Most methods relax the problem and only predict the coordinates of the body skeleton relative to a root joint, typically the hip \cite{3dbaseline,drpose,gorog}. In other words, the translation of the skeleton does not have to be calculated, only the limb lengths and orientations.

This may be sufficient if the image contains a single person only, as is the case with the popular human pose datasets like HumanEva \cite{humaneva} or Human3.6m \cite{h36m}. However, in videos containing interactions, the distance between actors and the environment can be important too.  For example, detecting hand-shakes, object manipulation and passing all require more information than the root-relative pose. To our knowledge, the only solution for absolute pose estimation is finding an optimal translation vector that minimizes the reprojection error \cite{mehta,zanfir2018smpl3dpose}. The search for the optimal translation is performed as a post-processing step after the root-relative 3D coordinates of the body joints have been determined.

This approach has several drawbacks: the relative 3D pose estimator is trained without the knowledge of the post-processing step. Thus, it misses information during back-propagation such as the size and distance of the person. Also, if the 3D pose estimator returns incorrect predictions, the translation vector that minimizes the reprojection error may diverge without limit (see Fig.~\ref{fig:baseline-err}). To overcome these issues, we propose a network that predicts absolute 3D coordinates instead of relative ones, circumventing the need for a translation optimization step. In our approach the origin of the coordinate system is the center of the camera.

Since absolute pose estimation is important in multi-people scenes and because studio videos have a very limited variance in depth, we use the MuPoTS-3D dataset \cite{mehta2018single_shot} that has multiple actors performing different activities in both outdoor and indoor settings. This dataset contains only evaluation data and no training data. Following \cite{mehta2018single_shot}, we have the MuCo-3DHP dataset as our training set. MuCo-3DHP  was introduced in the same paper \cite{mehta2018single_shot}.

However, MuCo-3DHP consists of synthesized studio scenes. To overcome the lack of variety in the training data, our network has a multi-stage architecture. The network first predicts 2D poses from the image and then predicts 3D coordinates using only the 2D output of the previous stage. Since large 2D annotated databases exist, the first step can be performed with high accuracy. This approach was successfully employed in a number of algorithms \cite{3dbaseline,fang2018posegrammar,lee2018pLSTM}, reaching state-of-the-art results. 

Since we would like to estimate absolute coordinates, not only root-relative ones, image details such as the relative position of people, the location of furniture, etc. might hold important information. To exploit those features without the need of a very large training set, we employ a depth estimation network. 

Depth predictor networks try to represent the scene geometry by predicting the depth for each pixel, essentially producing a 3D point-cloud from an input image \cite{monodepth2017,megadepth2018,fcrnd}. 
Although the predicted depth might not be accurate, for pose estimation predicting good ordinal ordering of joints (whether point A is closer or further from the camera than point B) already yields large improvements \cite{pavlakos2018ordinal,fbipose,drpose}. Birmingham et al. found that the results of MonoDepth \cite{monodepth2017} correlate  with human predictions of depth ranking \cite{brandon2018monodepthgoodrelative}. This motivates our choice to include a depth prediction network into our pipeline.

The depth predictor network can be used as a separate component, extracting features from the image that the 2D-to-3D network uses. This can be improved by training the network together with the 2D-to-3D network end-to-end.

To summarize, our contributions are as follows: 1) we introduce an architecture that predicts absolute 3D coordinates in one step, 2) we show that the addition of depth features provide significant performance increases and the depth network can be trained end-to-end with the pose estimating network, 3) we beat the previous state-of-the-art method on the MuPoTS-3D dataset. We make our code publicly available\footnote{\textbf{\texttt{https://github.com/vegesm/depthpose}}}.

\begin{figure*}[ht!]
\centering
\includegraphics[width=0.99\textwidth]{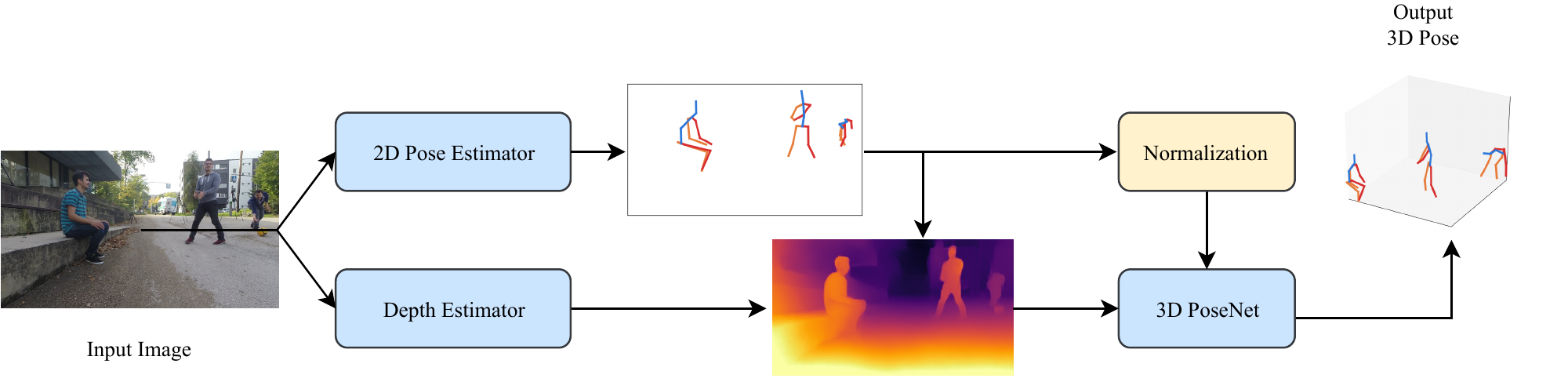}
\caption{Our network architecture. The 2D pose estimator detects body keypoints in the image while the depth estimator calculates the depth for each pixel. The depth is read out at the detected keypoint coordinates. The predicted 2D coordinates are further normalized with the inverse of the camera intrinsic matrix. The depth values and the focus normalized 2D coordinates are fed to the 3D PoseNet regressor that calculates the final 3D output.}
\label{fig:architecture}
\end{figure*}
In what follows, we treat related works (Sec.~\ref{sec:rel_works}) followed by the overview of our methods (Sect.~\ref{sec:methods}), the section about the experiments (Sect.~\ref{sec:experiments}) and our results (Sect.~\ref{sec:results}). We conclude in the last section (Sec.~\ref{sec:conc}).

\section{Related Work}\label{sec:rel_works}
\subsection{Relative pose estimation} Most 3D pose predictors estimate the body joint coordinates relative to a root joint, usually the hip. In \cite{tome2017liftingfromdeep} the authors propose a multi-stage architecture using a probabilistic model to predict depth coordinates. Pavlakos et al. \cite{gorog} predicts a 3D heatmap instead of single coordinates. Another set of methods use the soft-argmax function to go from a predicted 2D heatmap to 3D coordinates \cite{integralPose,Luvizon2018softargmax}. 

One problem with 3D prediction is the lack of varied in-the-wild datasets. Annotating an image with 3D information requires special equipment with multiple cameras. In contrast, 2D pose estimation datasets are easy to create so it seems beneficial to use them. One approach is to split the 3D estimation into two steps: first predict the 2D coordinates from the image and then predict 3D coordinates from the 2D coordinates only \cite{3dbaseline,fang2018posegrammar,lee2018pLSTM,veges2018siamese}. While it seems too limiting using just 2D coordinates and no other image features, these methods achieved state-of-the-art results nonetheless.

Another approach is based on the idea that humans are good at telling which of two points of an image is closer to the camera. Whereas it is nearly impossible for a person to guess distances in a photograph with high accuracy, annotating ordinal ranking of joints requires less than a minute \cite{pavlakos2018ordinal}. 2D datasets supplied with ordinal rankings can provide weak supervisory information \cite{pavlakos2018ordinal,fbipose,ronchi2018allrelative}. Even without annotating additional data, just predicting joint ranking information as an auxiliary task leads to improvements \cite{drpose}.

\subsection{Absolute pose estimation} While methods to estimate root-relative coordinates are numerous, only a handful predicts coordinates in a global system. Mehta et al. \cite{mehta} first predicts a 3D pose from an image and then finds an optimal translation minimizing the squared reprojection error. The least squares problem has an exact solution assuming weak projection.

In \cite{zanfir2018smpl3dpose}, the authors predict a full body mesh using the skinned multi-person linear (SMPL) model \cite{smpl2015}. First they predict an initial pose with the DMHS detector \cite{dmhs2017} and refine the prediction using multiple constraints, including reprojection error, a semantic loss involving body part segmentation and matching to ground plane. Similarly to \cite{mehta}, the positioning of the person in the global scene happens as a separate step. Our work is different to previous approaches in that the global pose estimation is performed directly.

\subsection{Depth estimation} We shortly review here recent approaches to depth estimation. Current methods all use some form of fully convolutional networks. Laina et al. \cite{fcrnd} uses a slightly modified ResNet\nobreakdash-50 with faster up-convolution blocks. In \cite{diw2016}, the authors use an HourGlass \cite{stacked_hourglass} like architecture, where convolutional filters were replaced with Inception-style modules \cite{inceptionnet}. Finally, MegaDepth \cite{megadepth2018} introduces a training set synthesized from images from the Internet, making learned models more robust.

Another branch of research uses consistency between different views to learn depth in an unsupervised manner. \mbox{MonoDepth} \cite{monodepth2017} takes as input a pair of images from stereo cameras and learns to reconstruct one view from the other via estimating depth. The method is very strong, even beating supervised algorithms.  In contrast to MonoDepth, DF-Net \cite{zou2018dfnet} does not need stereo images but consecutive frames from a monocular video. It predicts  optical flow and depth jointly, seeking consistency between successive frames. Note that during inference, both methods need only a single picture.

\section{Method}\label{sec:methods}

\begin{figure*}
\centering
\includegraphics{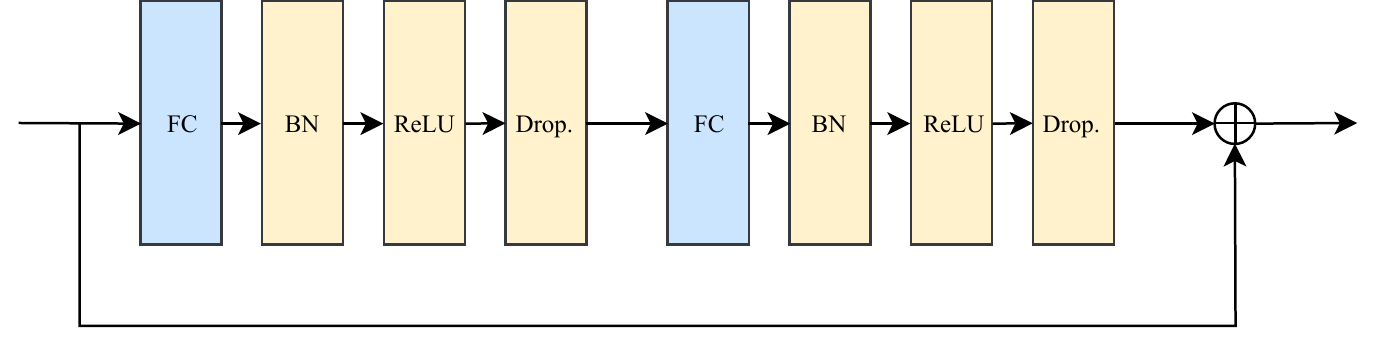}
\caption{One residual module in 3D PoseNet. It has two fully connected layers each followed by a Batch Normalization layer, a ReLU activation and a Dropout. The 3D PoseNet has two of these modules. The figure is taken from \cite{veges2018siamese}}.
\label{fig:res-block}
\end{figure*}

In this section we briefly describe the baseline method introduced in \cite{mehta} and also detail our network's architecture.

\subsection{Baseline}
The problem of finding global coordinates for a skeleton having the root-relative coordinates can be formulated as follows:
$$\hat{t}=\argmin_{t\in \mathbb{R}^3}\left\Vert P^{2D}-\Pi\left(P^{3D}+t\right)\right\Vert^2_2,$$
where $P^{2D}$ and $P^{3D}$ are the body joint coordinates in the image and 3D space, and $\Pi$ is the projection from 3D space to camera frame. The optimal translation $\hat{t}$ can be added to the root-relative pose $P^{3D}$ to get the final coordinates in a global coordinate system.

The minimization problem can be solved exactly, assuming a weak perspective projection:
$$\hat{t}=\alpha\left(\begin{matrix} \bar{P}^{2D}\\ f \end{matrix}\right)-\left(\begin{matrix} \bar{P}^{3D'}\\ 0 \end{matrix}\right),$$
$$\alpha=\frac{\sum_i\Vert P^{3D'}_i-\bar{P}^{3D'}\Vert^2_2}{\sum_i\langle P^{2D}_i-\bar{P}^{2D},P^{3D'}_i-\bar{P}^{3D'}\rangle},$$
where $P^{3D'}$ is the $x$ and $y$ coordinates from $P^{3D}$, $\bar{P}^{2D}$ and $\bar{P}^{3D'}$ are the means of  $P^{2D}$ and $P^{3D'}$. The derivation can be found in \cite{mehta}. Note that in their paper  the authors added an approximation at the last step which we do not include. Without the approximation we get better results.

In the full pipeline, the relative body pose is predicted using the same 2D pose estimator and 3D PoseNet that our method uses. To keep the results comparable, all details, including the network architecture, normalization and loss are the same in the baseline and our method. These are described in the next two sections.

\subsection{Our model}
The architecture of our network is sketched in Fig.~\ref{fig:architecture}. Based on the success of numerous earlier work (\cite{3dbaseline,fang2018posegrammar,lee2018pLSTM}),  we separate the 3D detection task into two steps: first we detect the 2D coordinates then regress the 3D coordinates from the 2D coordinates only. The 2D pose detector is the state-of-the-art multi-person pose detector OpenPose \cite{openpose}. In our network, the 2D-to-3D component is called \emph{3D PoseNet}.

To be robust against the change of cameras between the training and test set we normalize the 2D pose by multiplying with the inverse of the intrinsic camera matrix $K$. The normalized 2D pose is further processed by splitting the representation into two parts: the hip-relative coordinates of all the joints (except the hip as it is zero) and the original coordinates of the hip. This way the root-relative coordinates remain translation invariant. The disadvantage of this approach is that if the hip was not found by the 2D pose estimator then the entire person must be reported as undetected. However, we have found that an invisible hip implies a mostly invisible body in nearly all the cases. The number of poses thrown away because of this is just 3\% of all the frames. 

To make the 3D PoseNet able to use image features in addition to the 2D coordinates, we employ a depth estimator network. We could use an off-the-shelf depth estimator as a separate component, reading out the results at joint coordinates predicted by the 2D pose estimator.  The produced depth values act as additional features for the 3D PoseNet. This already leads to improvements as shown in Section \ref{sec:results}.

However, a direct readout have problems: the 2D pose estimator can predict an incorrect location that falls on the background, the joint can be occluded by another person or the depth prediction misses a limb. To overcome these issues, we train the depth estimator and the 3D PoseNet together, starting from a pretrained depth estimator. We chose the MegaDepth algorithm \cite{megadepth2018} as it was robust against the different indoor and outdoor settings in the MuPoTS-3D test set. In preliminary studies, other depth detectors trained  either on an indoor or outdoor database performed worse.

Since MegaDepth outputs the logarithm of the depth, we use log-depth as the input of 3D PoseNet. Also, to avoid the network having to learn an exponential function, the output hip depth is also given in a logarithmic scale. The predicted depth is not logarithmic for the other joints to keep translation invariance. 

The depth prediction and normalized coordinates are fed to the 3D PoseNet that generates the absolute 3D coordinates (Fig.~\ref{fig:architecture}). Similarly to the 2D poses, the output 3D pose is split into hip-relative and absolute part, helping the generalization ability of the network.

The architecture of the 3D PoseNet was inspired by \cite{3dbaseline} and is illustrated on Fig.~\ref{fig:res-block}. It consists of two blocks of residual modules, each having a dense layer followed by BatchNorm \cite{batchnorm} and Dropout \cite{dropout} layers. The activation function was ReLU.

\subsection{Training loss}
The loss function is calculated on the predicted coordinates for each person in the picture. We use the L1 loss to be robust against outliers. This is in line with the findings of previous work \cite{integralPose,sarandi2018eccv_winner}. The final loss is thus:
$$L=\frac{1}{N_P}\sum_{i=1}^{N_P}\left\vert\hat{P}^{3D}_i-P^{3D}_i\right\vert,$$
where $N_P$ is the number of detected poses in a batch. Note that $N_P$ changes across batches, as the number of people on an image varies.

\section{Experiments}\label{sec:experiments}

\subsection{Datasets}
We used the recently released MuPoTS-3D dataset \cite{mehta2018single_shot} for evaluation. Unlike other popular datasets, videos in MuPoTS-3D were not shot in a studio and have multiple people interacting. The database has 5 indoor and 15 outdoor scenes with buildings, trees and other objects. In total it has 8300 frames and 20k poses from 8 actors.

Since MuPoTS-3D does not have a training set, following \cite{mehta2018single_shot} we use the MuCo-3DHP dataset \cite{mehta2018single_shot} for training. The dataset contains synthetic images generated from the poses in the MPI-INF-3DHP database \cite{mehta}. Each picture in MuCo-3DHP is a composition of 4 frames from MPI-INF-3DHP from the same camera. The authors only provide the generating scripts for the dataset, not the images themselves. We created 150k training images with 4 people on each image. The script has an option for background augmentation, though it simply places an image behind the actors. This would interfere with the depth estimation and we chose not to use it.

Note that the training and test sets contain quite different pictures: the camera characteristics (resolution, focal length, position), scene backgrounds and the actors are all different. This ensures that the measured results are robust and not a product of overfitting.

\subsection{Evaluation metrics}

We employ a variation of the standard  mean per joint position error (MPJPE) metric to evaluate our models \cite{h36m}. The MPJPE metric is the root-relative Euclidean error averaged over all joints and poses. In a root-relative pose, the hip (the root joint) is set at the origin. Since we are interested in coordinates in a global space, we do not move the hip to the origin. We call the latter metric Absolute MPJPE or A-MPJPE for short. The original MPJPE is called Relative or R-MPJPE to avoid confusion.

In absolute pose estimation there could be two sources of errors: the (root-relative) pose is incorrectly estimated, or the absolute location of the pose is incorrect. The scale of the second type of error can be much larger then the first type. We report both metrics to avoid that the absolute error hides an inaccurate pose prediction. 

To summarize, the definition of the metrics:

\begin{itemize}
    \item \textbf{A-MPJPE} or Absolute MPJPE. The average Euclidean distance between the ground truth and predicted joints in millimeters.
    \item \textbf{R-MPJPE} or Relative MPJPE. The average Euclidean distance between the ground truth and predicted \textit{hip\nobreakdash-relative} joint coordinates in millimeters. Previous work calls this the MPJPE metric.
\end{itemize}
Thus, the A-MPJPE metric is a natural extension of the common MPJPE metric.

\subsection{Implementation details}
For the 2D Pose estimator we used the OpenPose \cite{openpose} multi-person pose estimator. It predicts 25 joints with a confidence score. We only use 14 joints as the rest is noisy and leads to degraded performance. The selected joints were: nose, neck, pelvis and left/right hip, knee, ankle, shoulder, elbow and wrist. Also note that annotations in the test set contains none of the 11 excluded joints. If OpenPose was unable to detect the hip the pose was discarded as undetected. We have found that the poses excluded this way were heavily occluded and hard to detect.

In the baseline algorithm, we used OpenPose with the 3D PoseNet together (using the same normalization techniques as described above). The 3D PoseNet had two blocks of residual modules depicted in Fig.~\ref{fig:res-block}, both module had two fully connected layers of 1024 neurons. The dropout rate was set to 0.5 as in \cite{3dbaseline}. We trained the network for 100 epochs with the Adam optimization algorithm. The learning rate was 0.001 initially and was decreased with a multiplier of 0.96 every 4 epochs. 

The training of our network was performed in two steps: first the depth estimator was fixed and only the 3D PoseNet was trained using Adam and a learning rate of 0.001 for 100 epochs. The learning rate was decreased the same way as with the baseline. The batch size was 256. In the second step the depth prediction network was trained as well. Since the network is much larger, only the top convolutional layer was updated, the rest remained fixed. Due to memory reasons, the batch size was decreased to 30 and the training ran for 5 epochs. Again, the Adam optimization algorithm was used with a learning rate of $10^{-5}$. During training we augmented the dataset by cropping and zooming the input images. This augmentation was used in the baseline experiments as well to have comparable results.

\section{Results}\label{sec:results}

\subsection{Pose estimation performance}
\begin{table}[htb]
\caption{Pose Estimation Performance on the MuPoTS-3D test set. Best result selected in bold. Errors are in mm. }
\begin{center}
\begin{tabular}{lccc}
 & A-MPJPE & R-MPJPE & Detection Rate \\
\hline
LCR-Net \cite{rogez2017lcrnet} & - & 146 & 86\% \\
Mehta et al. \cite{mehta2018single_shot} & - & 132 & \textbf{93\%}  \\
\hline
Baseline & 320 & 122 & 91\%  \\
Ours  & \textbf{292} & \textbf{120} & 91\%  \\
\hline
\end{tabular}
\label{tab:pose-perf}
\end{center}
\end{table}
In this section we review the absolute and relative pose estimation performance of our network compared to the baseline algorithm. Quantitative results are presented in Table~\ref{tab:pose-perf}. First note that in relative pose estimation (R-MPJPE metric), our baseline algorithm already beats the state-of-the-art on the MuPoTS-3D dataset by 10 mm (7.6\%), signaling the strength of the method.

Our end-to-end trained method achieved the best overall results both on the A-MPJPE and R-MPJPE metrics. In absolute pose estimation the improvement is 28mm (8.7\%). The relative error also decreased to 120mm (1.6\%). In the case of the relative pose estimation, the baseline and our method differ only in the depth features that we included. Thus the improvement of R-MPJPE comes from the depth estimator solely.

\begin{figure}
\centering
\includegraphics[width=0.95\linewidth]{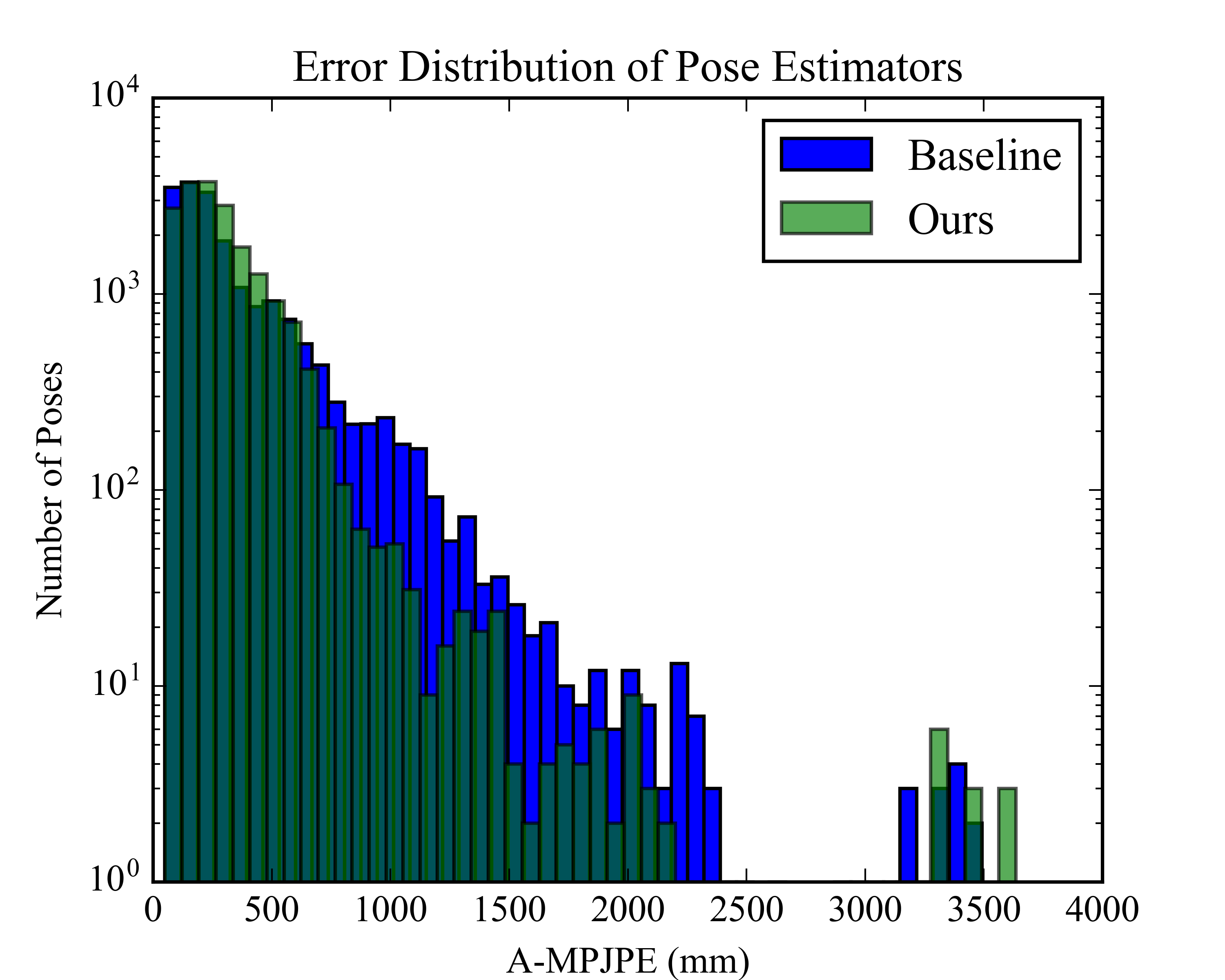}
\caption{Histogram of the error distributions of the baseline and our method (note the logarithmic scale for the number of poses). Our method has much fewer predictions with large error, particularly above 500mm}.
\label{fig:err_dist}
\end{figure}

\begin{figure}
\centering
\begin{tabular}{ccc}
Input & Baseline & Ours \\
\includegraphics[width = 0.3\linewidth]{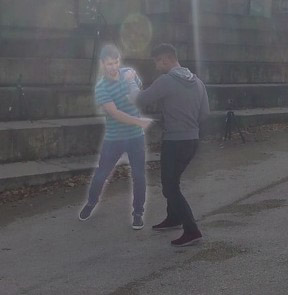} & 
\includegraphics[width = 0.3\linewidth]{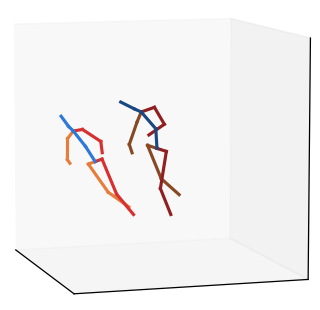} &
\includegraphics[width = 0.3\linewidth]{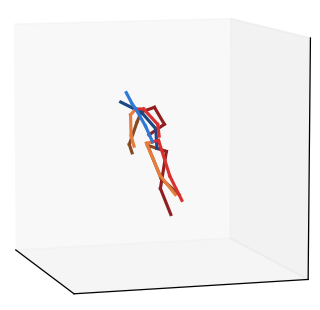}  \\
\end{tabular}
\caption{Typical error of baseline. For clarity, only one person is shown and from a different angle. Dark skeleton is the ground truth, light colored skeletons are the estimates. Since the predicted (root relative) 3D pose is incorrect, the baseline's reprojection error minimisation step places the pose far away from the ground truth. In contrast, due to the direct estimation of absolute coordinates, our model places the person at the correct position, even when the pose is incorrect. }
\label{fig:baseline-err}
\end{figure}

\begin{figure*}
\centering
\begin{tabular}{ccccccc}
Input & Baseline & Ours & & Input & Baseline & Ours \\
\includegraphics[width = 0.12\linewidth]{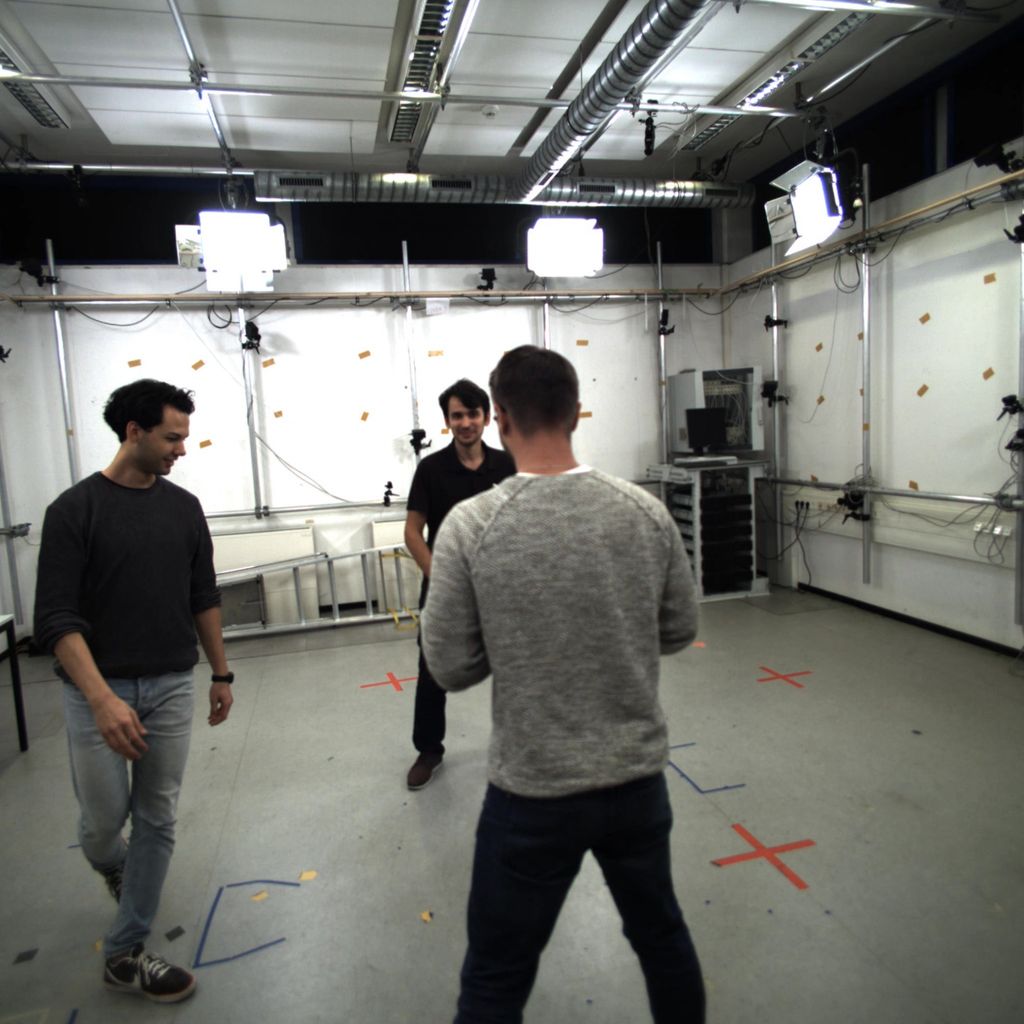} & 
\includegraphics[width = 0.12\linewidth]{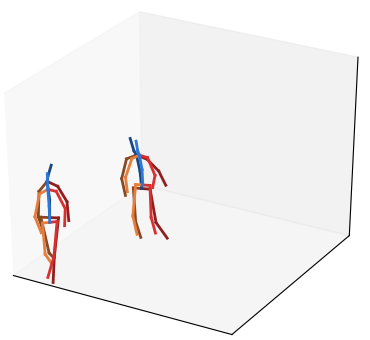} & 
\includegraphics[width = 0.12\linewidth]{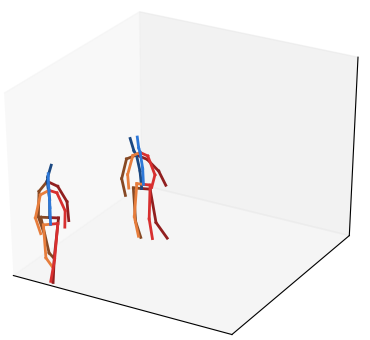} & 
\hspace{4em} &
\includegraphics[width = 0.12\linewidth]{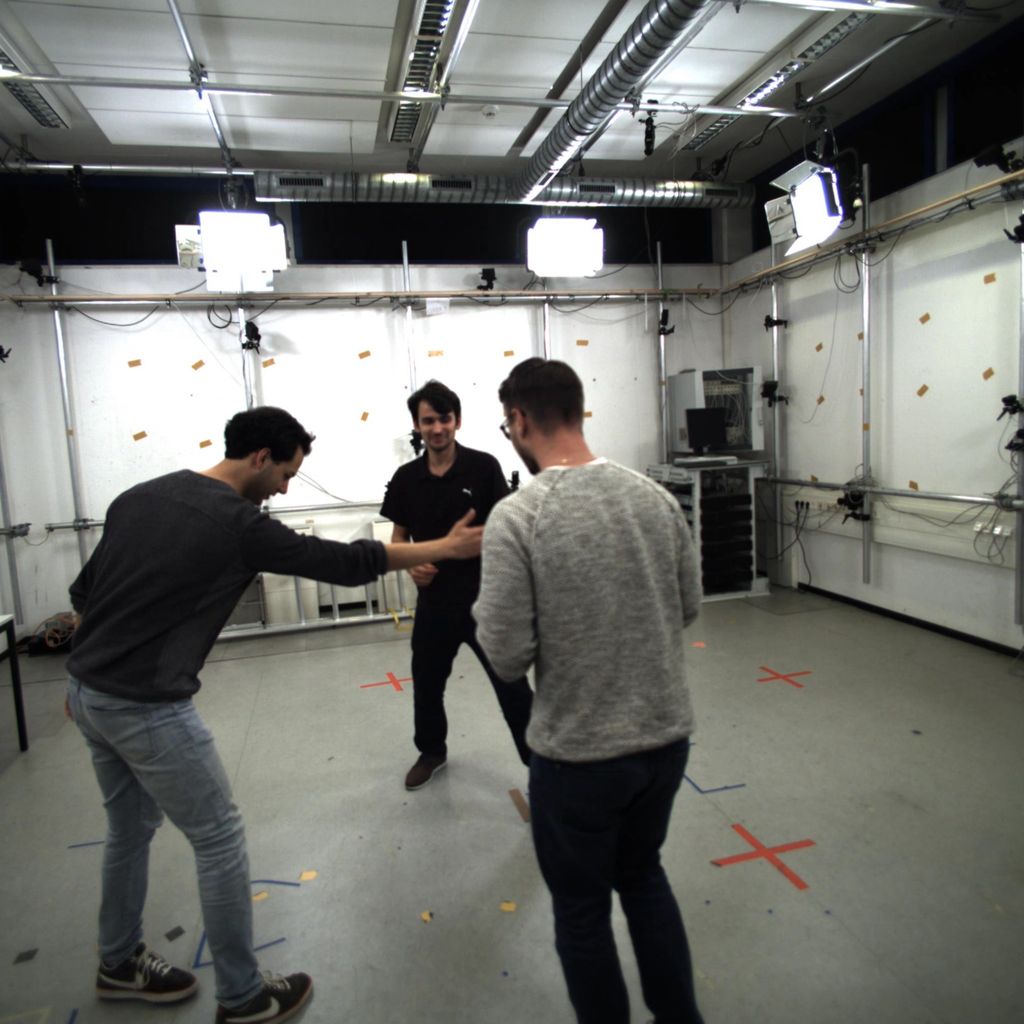} & 
\includegraphics[width = 0.12\linewidth]{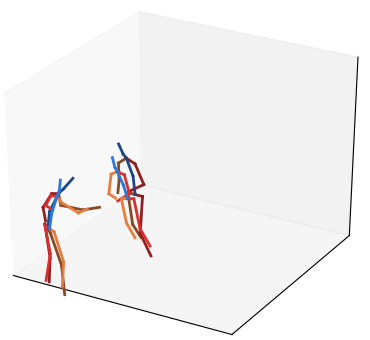} & 
\includegraphics[width = 0.12\linewidth]{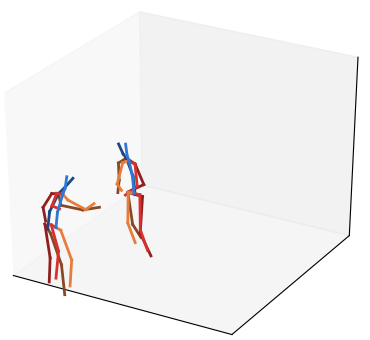} \\
\includegraphics[width = 0.12\linewidth]{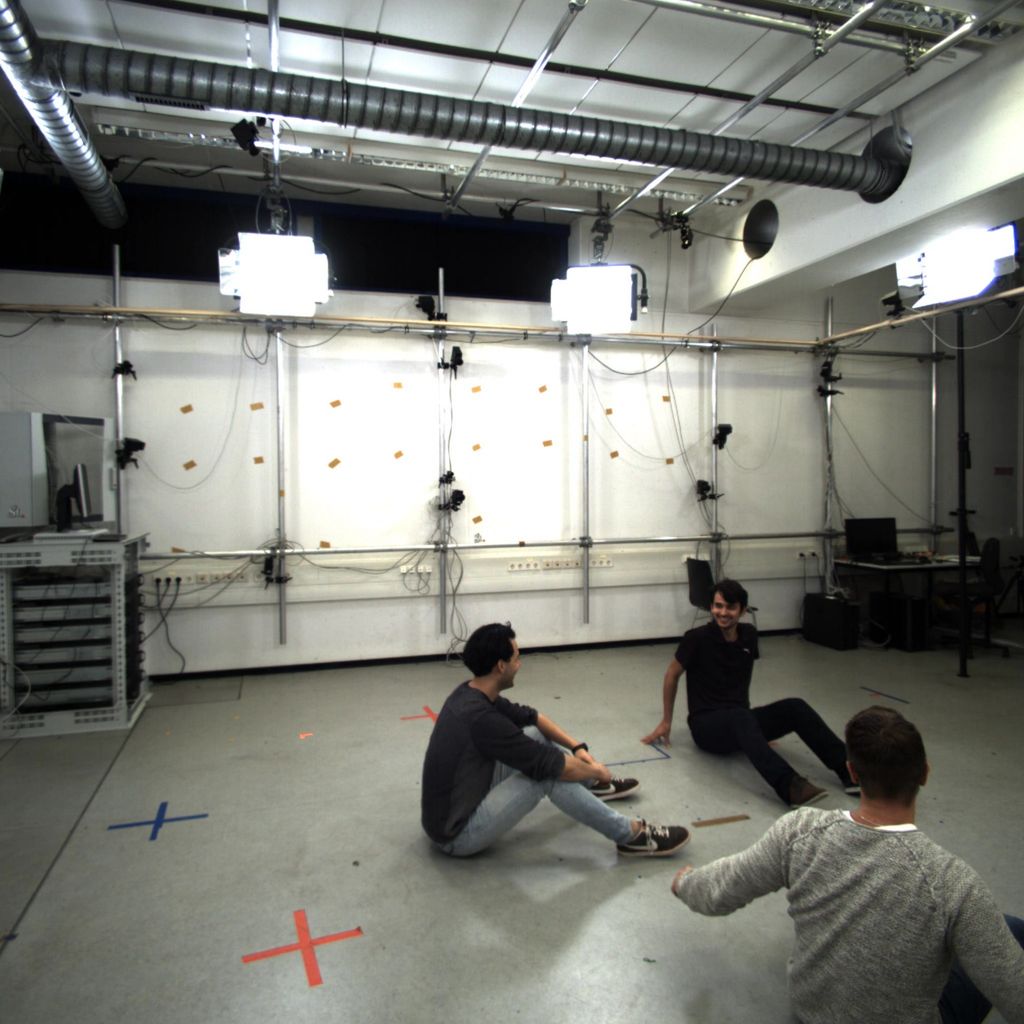} & 
\includegraphics[width = 0.12\linewidth]{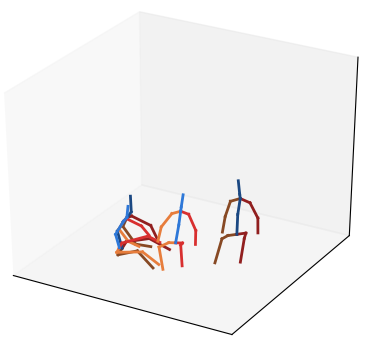} & 
\includegraphics[width = 0.12\linewidth]{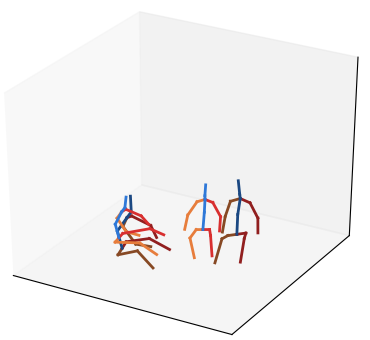} & 
\hspace{4em} &
\includegraphics[width = 0.12\linewidth]{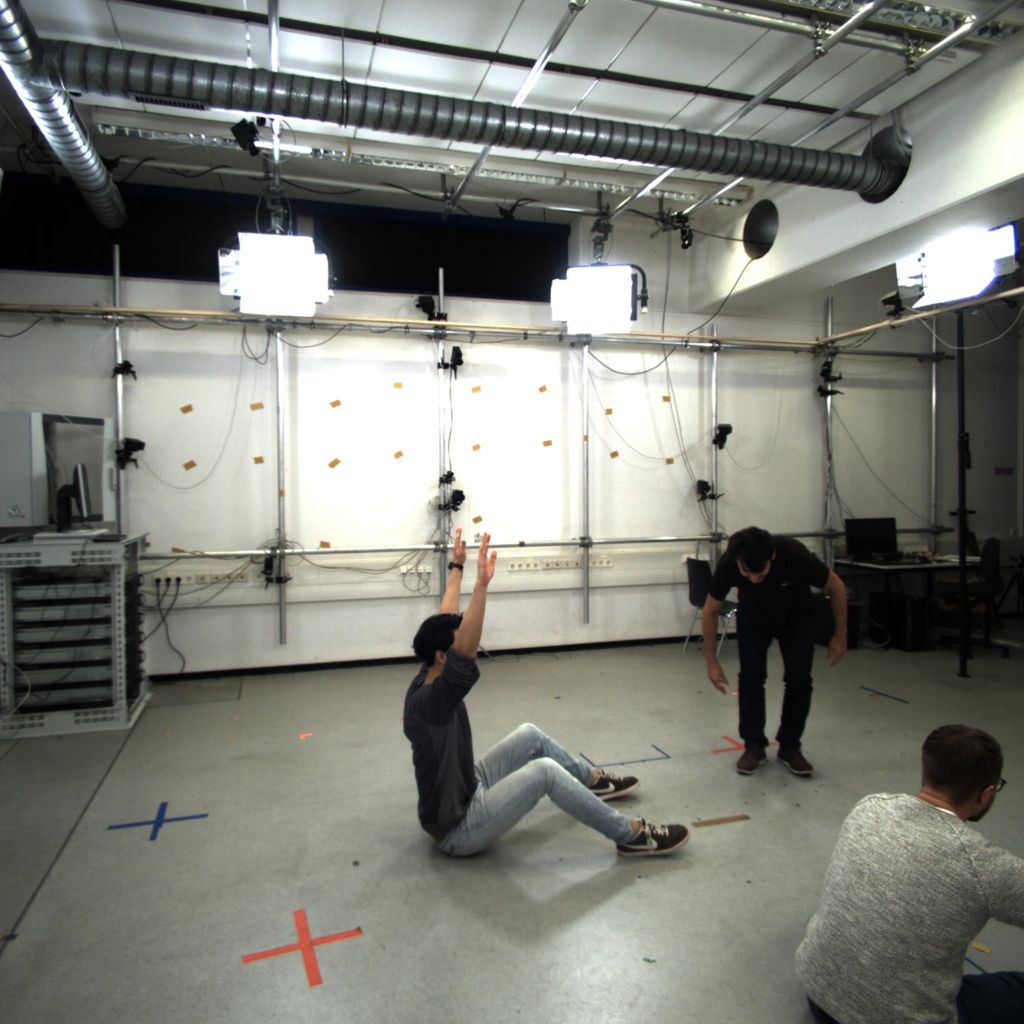} & 
\includegraphics[width = 0.12\linewidth]{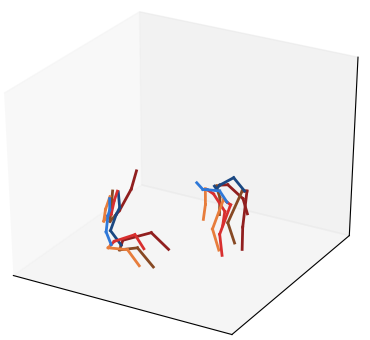} & 
\includegraphics[width = 0.12\linewidth]{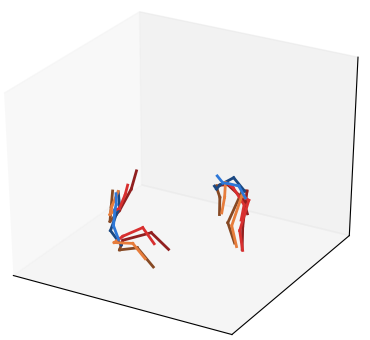} \\
\includegraphics[width = 0.12\linewidth]{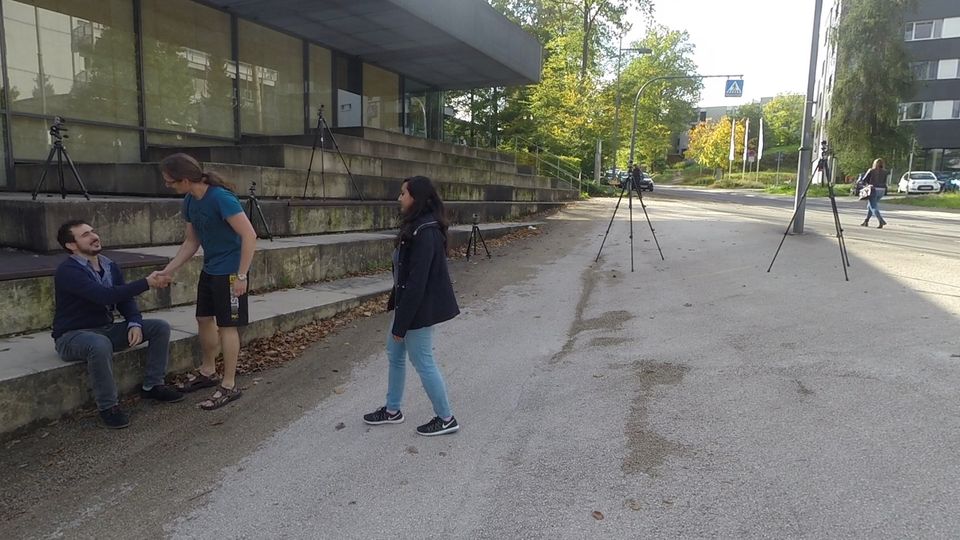} & 
\includegraphics[width = 0.12\linewidth]{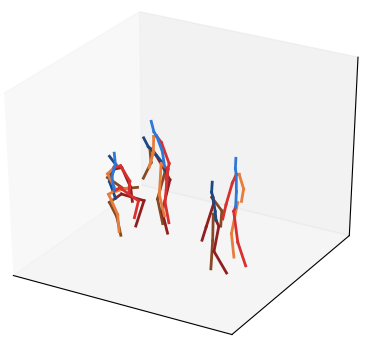} & 
\includegraphics[width = 0.12\linewidth]{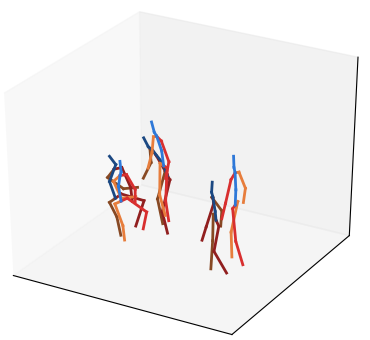} & 
\hspace{4em} &
\includegraphics[width = 0.12\linewidth]{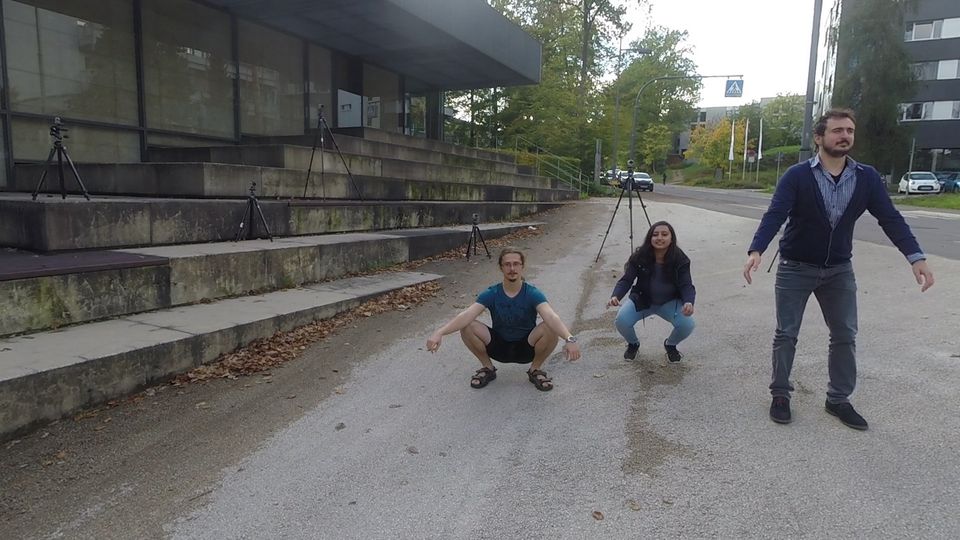} & 
\includegraphics[width = 0.12\linewidth]{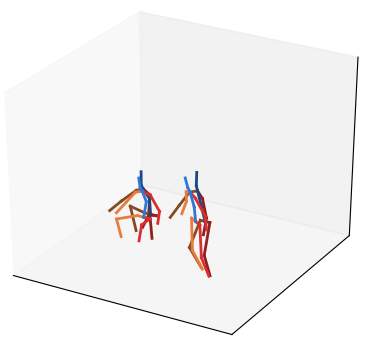} & 
\includegraphics[width = 0.12\linewidth]{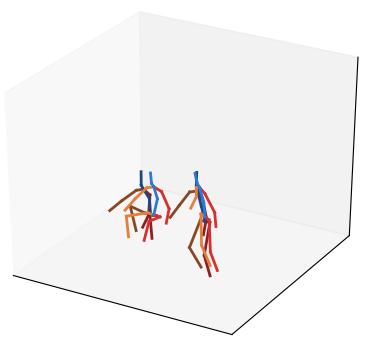} \\
\includegraphics[width = 0.12\linewidth]{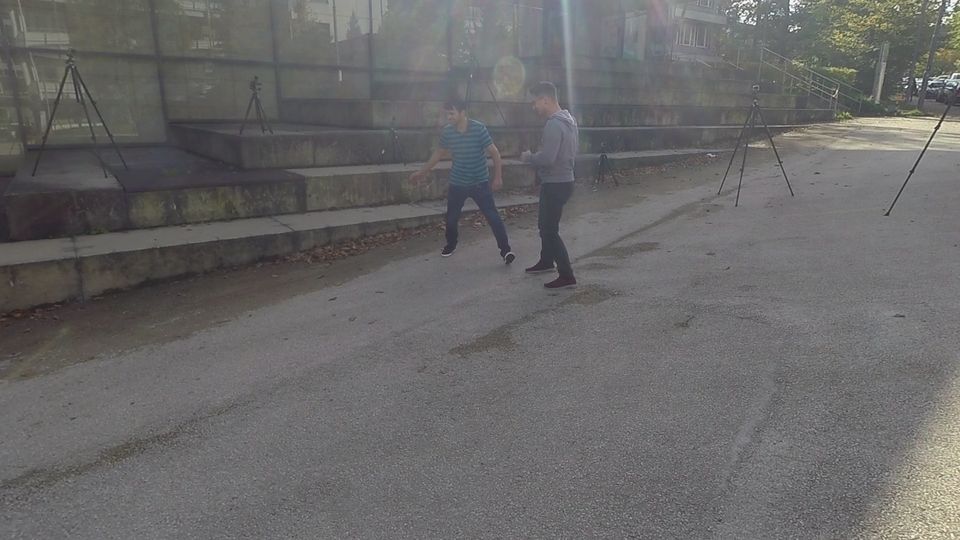} & 
\includegraphics[width = 0.12\linewidth]{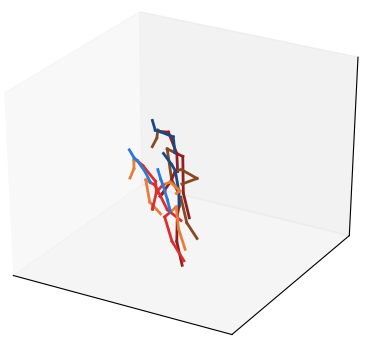} & 
\includegraphics[width = 0.12\linewidth]{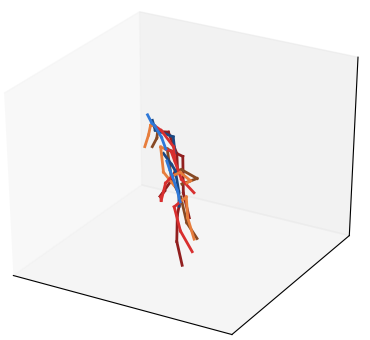} & 
\hspace{4em} &
\includegraphics[width = 0.12\linewidth]{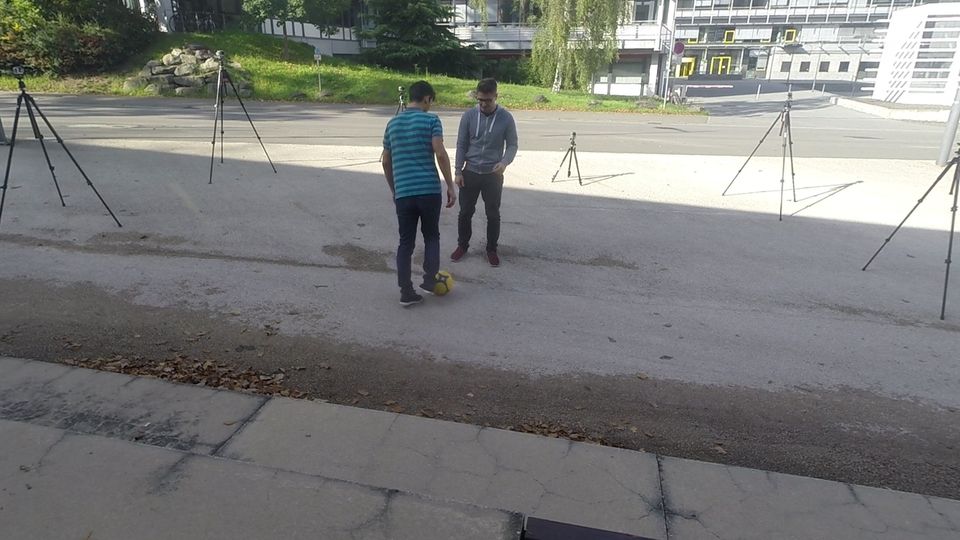} & 
\includegraphics[width = 0.12\linewidth]{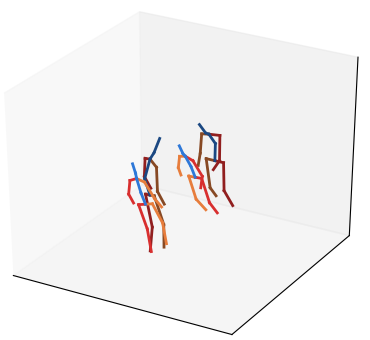} & 
\includegraphics[width = 0.12\linewidth]{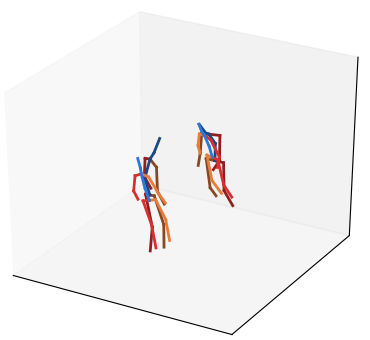} \\
\includegraphics[width = 0.12\linewidth]{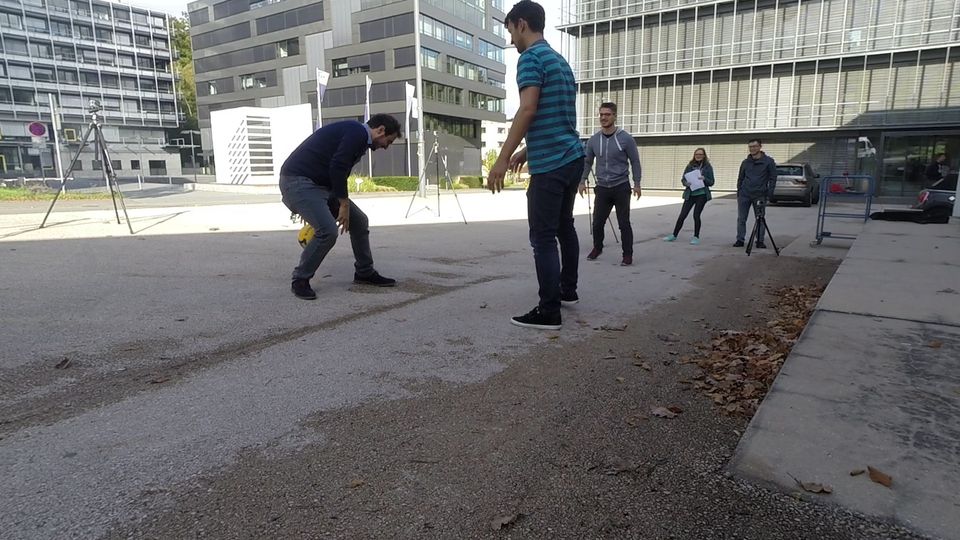} & 
\includegraphics[width = 0.12\linewidth]{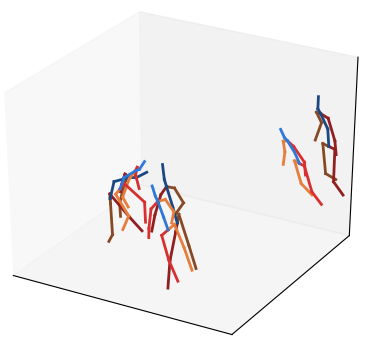} & 
\includegraphics[width = 0.12\linewidth]{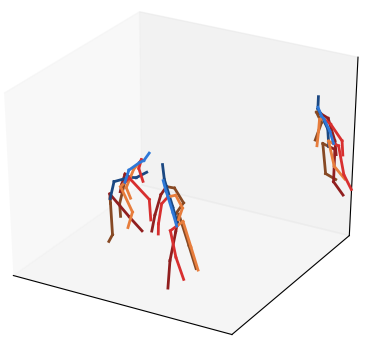} & 
\hspace{4em} &
\includegraphics[width = 0.12\linewidth]{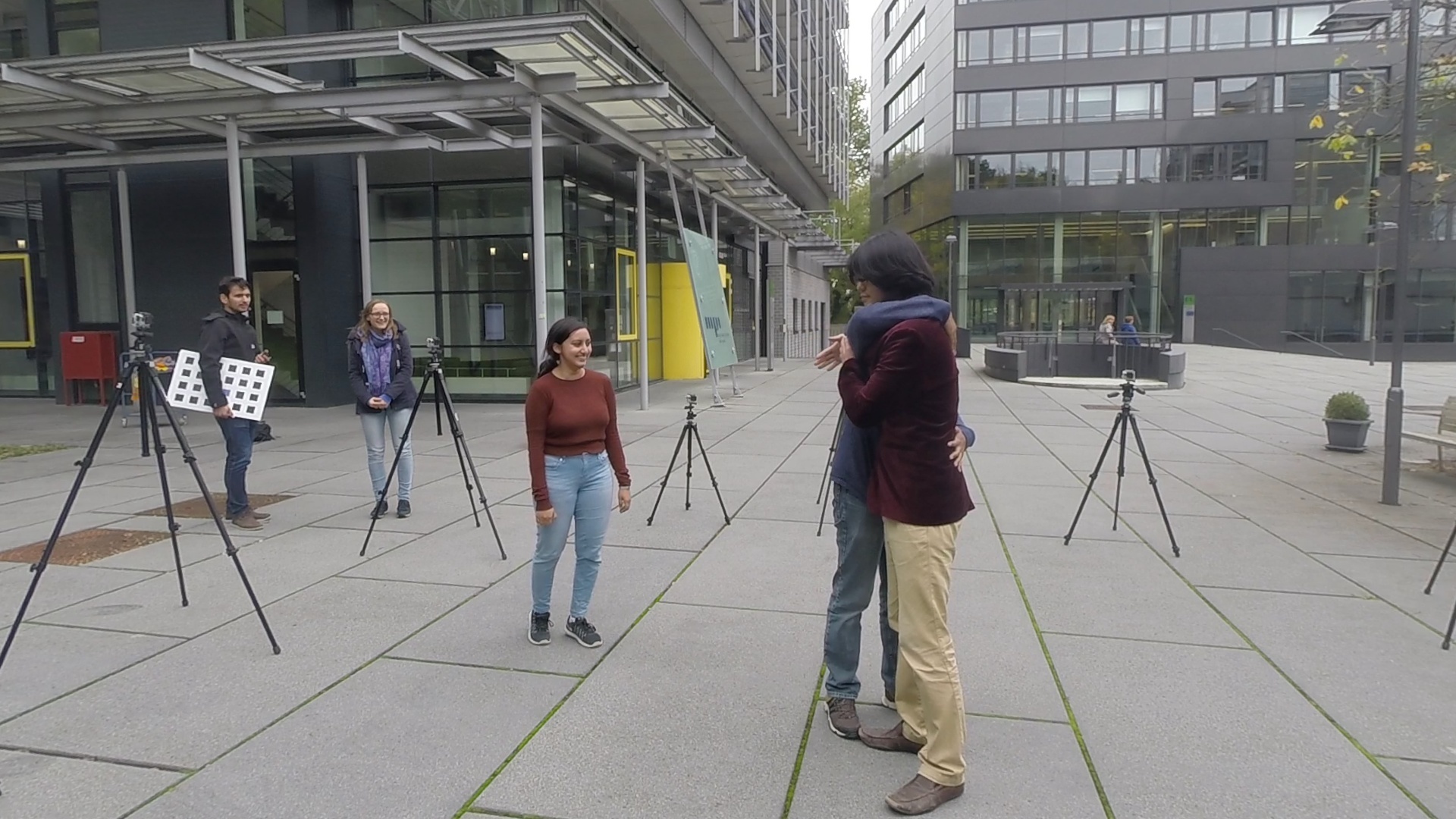} & 
\includegraphics[width = 0.12\linewidth]{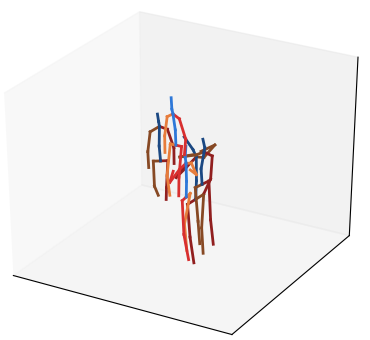} & 
\includegraphics[width = 0.12\linewidth]{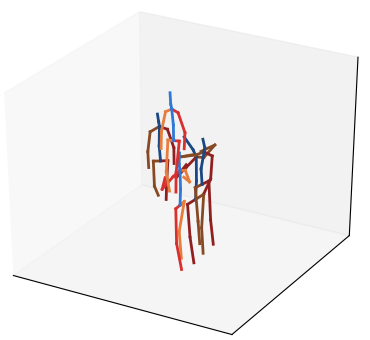} \\
\includegraphics[width = 0.12\linewidth]{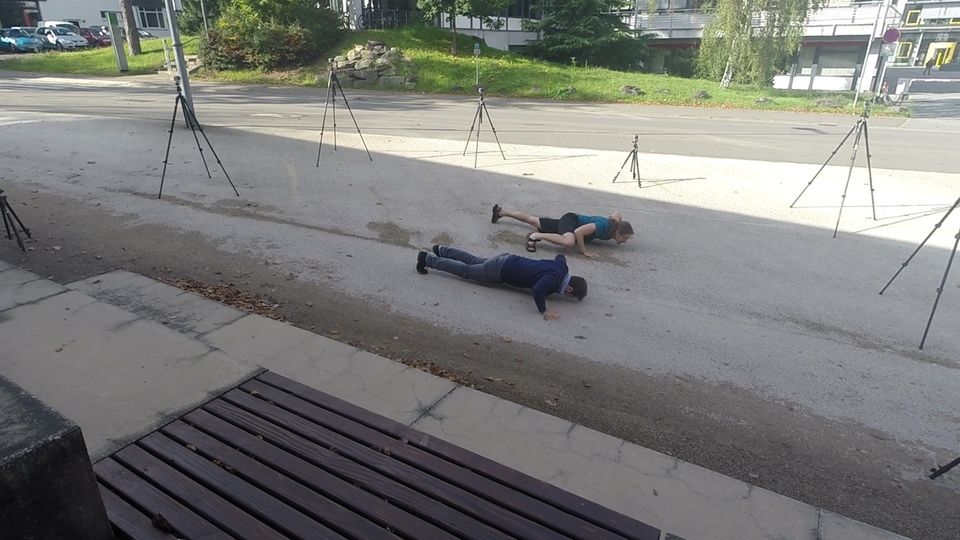} & 
\includegraphics[width = 0.12\linewidth]{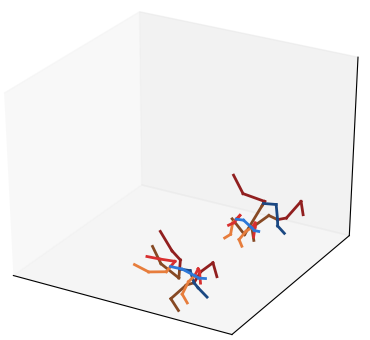} & 
\includegraphics[width = 0.12\linewidth]{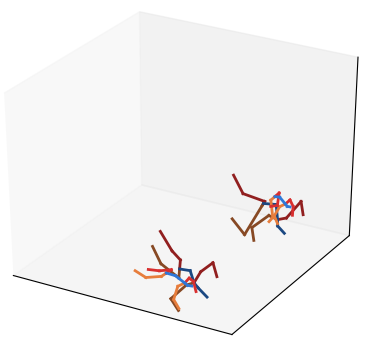} & 
\hspace{4em} &
\includegraphics[width = 0.12\linewidth]{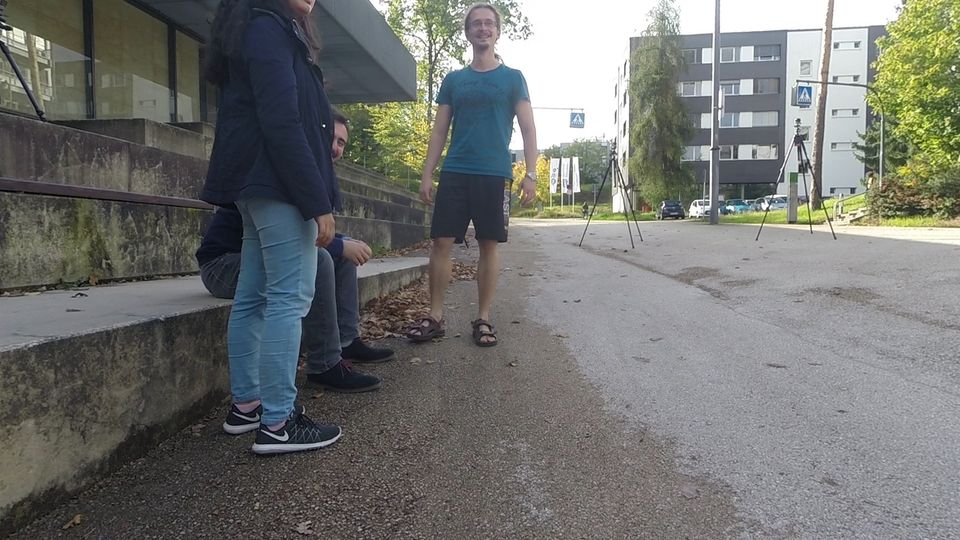} & 
\includegraphics[width = 0.12\linewidth]{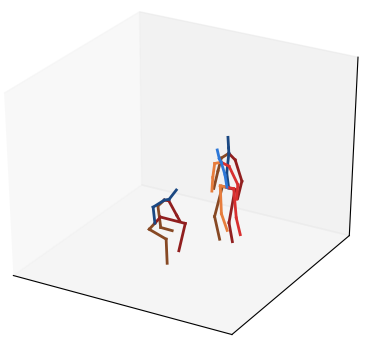} & 
\includegraphics[width = 0.12\linewidth]{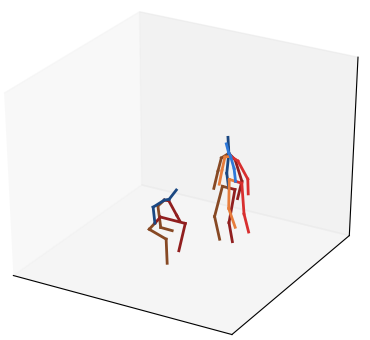} \\
\\

\end{tabular}
\caption{Qualitative results. Dark skeletons are ground truth values, light ones are the network predictions. Bottom left: an erroneous result, bottom right: a case when the hip is hidden. For more information, see text. Note: viewing angles differ from those of the images for visualization purposes. Also, not all people visible in the scene have ground-truth annotations, those are not displayed in the figure.}
\label{fig:qualitative-res}
\end{figure*}

To gain further insights on how the baseline and our method compares we present a histogram of the error distributions in Fig.~\ref{fig:err_dist}. Our method produces fewer large errors, as the shorter tail of the graph indicates. On the other hand, the baseline algorithm performs a bit better on the low-error range; it has more input samples where the prediction error was under 100mm. However, this does not compensate for the larger errors produced on other poses.

Our algorithm detects slightly less (2\%) poses than that of Mehta et al. \cite{mehta2018single_shot} but 5\% more than LCR-Net \cite{rogez2017lcrnet}. However, the baseline and our method have the same performance in this regard as the difference is only in the pose estimation part and not in the detection part.

To summarize, our end-to-end trained method achieves new state of the art relative pose estimation results on the MuPoTS-3D dataset and improves on the commonly used baseline method.

\subsection{Qualitative results}
We present sample outputs of our network on Fig.~\ref{fig:qualitative-res}. One can see that the relative pose is similar to the baseline's output. On the other hand, the skeletons are placed closer to the ground truth by our model. If two people are very close to each other (e.g. hugging), the 2D detector often fails to find one of the persons due to heavy occlusion, see Fig.~\ref{fig:qualitative-res} fifth row second column. Here one of the subjects is nearly fully occluded.

The left column of the bottom row shows a failure case where the detected (relative) pose is wrong. Note that our method still places the skeleton closer to the ground truth, while the baseline is unable to do that.

The right column of the bottom row shows an example where a person was undetected due to hidden hip. Most of the sitting person is occluded and it would be hard to make a good prediction of his pose.

Additionally, Fig.~\ref{fig:baseline-err} shows how our direct estimation solves a common problem of the two-step approach. The figure shows an input with a difficult pose for which both our and the baseline method returns an incorrect root-relative estimation. In the baseline method, not only the root-relative pose is faulty but the location of that pose is erroneous as well. This is due to the fact that no 2D reprojection of a bad 3D pose is close to the original detected 2D pose. However, our model correctly places the person in the space. In other words, the incorrect pose prediction does not prevent finding the correct location of the person.

\subsection{Ablation study}

\begin{table}[htb]
\caption{Ablation study results. The table shows the change in A-MPJPE (units in mm) while turning on  components of our network sequentially.}
\begin{center}
\begin{tabular}{lc}
 & A-MPJPE \\
\hline
L2 loss  &  421 \\
w/ L1 loss &  379 \\
w/ Depth features & 367 \\
Predicting log of hip z coordinate & 358 \\
Augmentation & 314 \\
End-to-end training & 292 \\
\hline
\end{tabular}
\label{tab:ablation}
\end{center}
\end{table}

We validate our design decisions by an ablation study. The results are presented in Table \ref{tab:ablation}. Changing from L2 loss to L1 improves the performance by 42~mm (10\%). The large drop can be attributed to the robustness of L1 against outliers. The inclusion of depth features from MegaDepth further decreases the error by 12~mm. Additional improvements can be achieved by predicting the logarithm of the $z$ coordinate of the hip. This choice was motivated by two facts: first, MegaDepth predicts logarithms of depth values, second, the distribution of depth coordinates has a long-tail distribution and the logarithm function essentially converts it back to a more symmetric one \cite{megadepth2018}.

Augmenting the data with crops and zoom leads to another significant drop of 44mm or 12\%. Finally, the end-to-end finetuning improves the results by an additional 22mm (7\%).

\section{Conclusions}\label{sec:conc}
We have introduced a single-step solution for absolute pose estimation on multi-person scenes. Unlike previous approaches, this does not require any post-processing. We also showed that depth estimators are good source of additional features for absolute pose estimation. This results in improved performance and state-of-the-art results on the MuPoTS-3D dataset. The dataset is different from the training set so it indicates a good generalization ability. 

Although the decrease in error metrics is significant, there is still space for further improvements. The 2D Pose Estimator can be trained end-to-end, together with the 3D PoseNet and depth estimator networks. Also, for many applications, such as detecting interactions between the subjects, estimates are fine to be given scale independently, as long as all the persons on an image are represented in the same scale. Thus instead of predicting absolute coordinates, one could estimate scale invariant ones. This removes the ambiguity from the problem statement.

\bibliographystyle{ieeetr}
\bibliography{refs}

\begin{thebibliography}{10}

\bibitem{openpose}
Z.~Cao, T.~Simon, S.-E. Wei, and Y.~Sheikh, ``Realtime multi-person 2d pose
  estimation using part affinity fields,'' in {\em The IEEE Conference on
  Computer Vision and Pattern Recognition}, pp.~1302--1310, 2017.

\bibitem{stacked_hourglass}
A.~Newell, K.~Yang, and J.~Deng, ``Stacked hourglass networks for human pose
  estimation,'' in {\em European Conference on Computer Vision}, pp.~483--499,
  2016.

\bibitem{alphapose}
H.-S. Fang, S.~Xie, Y.-W. Tai, and C.~Lu, ``Rmpe: Regional multi-person pose
  estimation,'' in {\em The IEEE International Conference on Computer Vision},
  pp.~2353--2362, 2017.

\bibitem{3dbaseline}
J.~Martinez, R.~Hossain, J.~Romero, and J.~J. Little, ``A simple yet effective
  baseline for 3d human pose estimation,'' in {\em The IEEE International
  Conference on Computer Vision}, pp.~2659--2668, 2017.

\bibitem{drpose}
M.~Wang, X.~Chen, W.~Liu, C.~Qian, L.~Lin, and L.~Ma, ``Drpose3d: Depth ranking
  in 3d human pose estimation,'' in {\em Proceedings of the Twenty-Seventh
  International Joint Conference on Artificial Intelligen}, pp.~978--984, 2018.

\bibitem{gorog}
G.~Pavlakos, X.~Zhou, K.~G. Derpanis, and K.~Daniilidis, ``Coarse-to-fine
  volumetric prediction for single-image 3d human pose,'' in {\em The IEEE
  Conference on Computer Vision and Pattern Recognition}, pp.~1263--1272, IEEE,
  2017.

\bibitem{humaneva}
L.~Sigal, A.~O. Balan, and M.~J. Black, ``Humaneva: Synchronized video and
  motion capture dataset and baseline algorithm for evaluation of articulated
  human motion,'' {\em International Journal of Computer Vision}, vol.~87,
  p.~4, Aug 2009.

\bibitem{h36m}
C.~Ionescu, D.~Papava, V.~Olaru, and C.~Sminchisescu, ``Human3.6m: Large scale
  datasets and predictive methods for 3d human sensing in natural
  environments,'' {\em IEEE Transactions on Pattern Analysis and Machine
  Intelligence}, vol.~36, pp.~1325--1339, jul 2014.

\bibitem{mehta}
D.~Mehta, H.~Rhodin, D.~Casas, P.~Fua, O.~Sotnychenko, W.~Xu, and C.~Theobalt,
  ``Monocular 3d human pose estimation in the wild using improved cnn
  supervision,'' in {\em International Conference on 3D Vision}, pp.~506--516,
  2017.

\bibitem{zanfir2018smpl3dpose}
A.~Zanfir, E.~Marinoiu, and C.~Sminchisescu, ``Monocular 3d pose and shape
  estimation of multiple people in natural scenes the importance of multiple
  scene constraints,'' {\em The IEEE Conference on Computer Vision and Pattern
  Recognition}, pp.~2148--2157, 2018.

\bibitem{mehta2018single_shot}
D.~Mehta, O.~Sotnychenko, F.~Mueller, W.~Xu, S.~Sridhar, G.~Pons-Moll, and
  C.~Theobalt, ``Single-shot multi-person 3d pose estimation from monocular
  rgb,'' in {\em International Conference on 3D Vision}, IEEE, sep 2018.

\bibitem{fang2018posegrammar}
H.-S. Fang, Y.~Xu, W.~Wang, X.~Liu, and S.-C. Zhu, ``Learning pose grammar to
  encode human body configuration for 3d pose estimation,'' {\em The AAAI
  Conference on Artificial Intelligence}, 2018.

\bibitem{lee2018pLSTM}
K.~Lee, I.~Lee, and S.~Lee, ``Propagating lstm: 3d pose estimation based on
  joint interdependency,'' in {\em The European Conference on Computer Vision},
  September 2018.

\bibitem{monodepth2017}
C.~Godard, O.~{Mac Aodha}, and G.~J. Brostow, ``Unsupervised monocular depth
  estimation with left-right consistency,'' in {\em The IEEE Conference on
  Computer Vision and Pattern Recognition}, 2017.

\bibitem{megadepth2018}
Z.~Li and N.~Snavely, ``Megadepth: Learning single-view depth prediction from
  internet photos,'' in {\em The IEEE Conference on Computer Vision and Pattern
  Recognition}, 2018.

\bibitem{fcrnd}
I.~Laina, C.~Rupprecht, V.~Belagiannis, F.~Tombari, and N.~Navab, ``Deeper
  depth prediction with fully convolutional residual networks,'' in {\em The
  Fourth International Conference on 3D Vision}, pp.~239--248, 2016.

\bibitem{pavlakos2018ordinal}
G.~Pavlakos, X.~Zhou, and K.~Daniilidis, ``Ordinal depth supervision for 3d
  human pose estimation,'' in {\em The IEEE Conference on Computer Vision and
  Pattern Recognition}, 2018.

\bibitem{fbipose}
Y.~Shi, X.~Han, N.~Jiang, K.~Zhou, K.~Jia, and J.~Lu, ``Fbi-pose: Towards
  bridging the gap between 2d images and 3d human poses using
  forward-or-backward information,'' in {\em unpublished}, 2018.
\newblock arXiv:1806.09241.

\bibitem{brandon2018monodepthgoodrelative}
B.~Birmingham, A.~Muscat, and A.~Belz, ``Adding the third dimension to spatial
  relation detection in 2d images,'' in {\em Proceedings of the 11th
  International Conference on Natural Language Generation}, pp.~146--151,
  Association for Computational Linguistics, 2018.

\bibitem{tome2017liftingfromdeep}
D.~Tome, C.~Russell, and L.~Agapito, ``Lifting from the deep: Convolutional 3d
  pose estimation from a single image,'' in {\em The IEEE Conference on
  Computer Vision and Pattern Recognition}, July 2017.

\bibitem{integralPose}
X.~Sun, B.~Xiao, S.~Liang, and Y.~Wei, ``Integral human pose regression,'' in
  {\em The European Conference on Computer Vision}, pp.~529--545, September
  2018.

\bibitem{Luvizon2018softargmax}
D.~C. Luvizon, D.~Picard, and H.~Tabia, ``2d/3d pose estimation and action
  recognition using multitask deep learning,'' in {\em The IEEE Conference on
  Computer Vision and Pattern Recognition}, 2018.

\bibitem{veges2018siamese}
M.~V\'eges, V.~Varga, and A.~L\H{o}rincz, ``3d human pose estimation with
  siamese equivariant embedding,'' {\em Neurocomputing, in print}, 2018.

\bibitem{ronchi2018allrelative}
M.~R. Ronchi, O.~{Mac Aodha}, R.~Eng, and P.~Perona, ``It's all relative:
  Monocular 3d human pose estimation from weakly supervised data,'' in {\em
  Proceedings of the British Machine Vision Conference}, 2018.

\bibitem{smpl2015}
M.~Loper, N.~Mahmood, J.~Romero, G.~Pons-Moll, and M.~J. Black, ``{SMPL}: A
  skinned multi-person linear model,'' {\em ACM Trans. Graphics (Proc. SIGGRAPH
  Asia)}, vol.~34, pp.~248:1--248:16, Oct. 2015.

\bibitem{dmhs2017}
A.-I. Popa, M.~Zanfir, and C.~Sminchisescu, ``Deep multitask architecture for
  integrated 2d and 3d human sensing,'' in {\em IEEE International Conference
  on Computer Vision and Pattern Recognition}, 2017.

\bibitem{diw2016}
W.~Chen, Z.~Fu, D.~Yang, and J.~Deng, ``Single-image depth perception in the
  wild,'' in {\em Advances in Neural Information Processing Systems 29} (D.~D.
  Lee, M.~Sugiyama, U.~V. Luxburg, I.~Guyon, and R.~Garnett, eds.),
  pp.~730--738, Curran Associates, Inc., 2016.

\bibitem{inceptionnet}
C.~Szegedy, W.~Liu, Y.~Jia, P.~Sermanet, S.~Reed, D.~Anguelov, D.~Erhan,
  V.~Vanhoucke, and A.~Rabinovich, ``Going deeper with convolutions,'' in {\em
  The IEEE Conference on Computer Vision and Pattern Recognition}, June 2015.

\bibitem{zou2018dfnet}
Y.~Zou, Z.~Luo, and J.-B. Huang, ``Df-net: Unsupervised joint learning of depth
  and flow using cross-task consistency,'' in {\em European Conference on
  Computer Vision}, 2018.

\bibitem{batchnorm}
S.~Ioffe and C.~Szegedy, ``Batch normalization: Accelerating deep network
  training by reducing internal covariate shift,'' in {\em International
  Conference on Machine Learning}, pp.~448--456, 2015.

\bibitem{dropout}
N.~Srivastava, G.~Hinton, A.~Krizhevsky, I.~Sutskever, and R.~Salakhutdinov,
  ``Dropout: A simple way to prevent neural networks from overfitting,'' {\em
  The Journal of Machine Learning Research}, vol.~15, no.~1, pp.~1929--1958,
  2014.

\bibitem{sarandi2018eccv_winner}
I.~Sarandi, T.~Linder, K.~O.~Arras, and B.~Leibe, ``Synthetic occlusion
  augmentation with volumetric heatmaps for the 2018 eccv posetrack challenge
  on 3d human pose estimation,'' in {\em unpublished}, 2018.
\newblock arXiv:1809.04987.

\bibitem{rogez2017lcrnet}
G.~Rogez, P.~Weinzaepfel, and C.~Schmid, ``Lcr-net:
  Localization-classification-regression for human pose,'' in {\em The IEEE
  Conference on Computer Vision and Pattern Recognition}, pp.~1216--1224, July
  2017.

\end{thebibliography}

\end{document}